\documentclass[9pt]{article}

\usepackage{multicol}
\usepackage{times}
\usepackage{epsfig}
\usepackage{graphicx}
\usepackage{amsmath}
\usepackage{caption}
\usepackage{helvet}
\usepackage{courier}
\usepackage{subcaption}
\usepackage{gensymb}
\usepackage{multirow}
\usepackage{textcomp}
\usepackage{cite}
\usepackage{algorithm}
\usepackage{algorithmic}

\title{Deep network for rolling shutter rectification}
\begin{document}
\date{}

\author{Praveen K, Lokesh Kumar T,and A.N. Rajagopalan }




\maketitle
\begin{abstract}
   CMOS sensors employ row-wise acquisition mechanism while imaging a scene, which can result in undesired motion artifacts known as rolling shutter (RS) distortions in the captured image. Existing single image RS rectification methods attempt to account for these distortions by either using algorithms tailored for specific class of scenes which warrants information of intrinsic camera parameters or a learning-based framework with known ground truth motion parameters. In this paper, we propose an end-to-end deep neural network for the challenging task of single image RS rectification. Our network consists of a motion block, a trajectory module, a row block, an RS rectification module and an RS regeneration module (which is used only during training). The motion block predicts camera pose for every row of the input RS distorted image while the trajectory module fits estimated motion parameters to a third-order polynomial. The row block predicts the camera motion that must be associated with every pixel in the target i.e, RS rectified image. Finally, the RS rectification module uses motion trajectory and the output of row block to warp the input RS image to arrive at a distortion-free image. For faster convergence during training, we additionally use an RS regeneration module which compares the input RS image with the ground truth image distorted by estimated motion parameters. The end-to-end formulation in our model does not constrain the estimated motion to ground-truth motion parameters, thereby successfully rectifying the RS images with complex real-life camera motion. Experiments on synthetic and real datasets reveal that our network outperforms prior art both qualitatively and quantitatively. 
\end{abstract}

\section{Introduction}
Most present-day cameras are equipped with CMOS sensors due to advantages such as slimmer readout circuitry, lower cost, and higher frame rate over their CCD counterparts.  While capturing an image, CMOS sensor array is exposed to a scene in a sequentially row-wise manner. The flip side is that, in the presence of camera motion, the inter-row delay leads to undesirable geometric effects, also known as rolling shutter (RS) distortions. This is because rows of the RS image do not necessarily sense the same camera motion. A prominent effect of RS distortion is the manifestation of straight lines as curves which call for correction of RS effect also termed as RS rectification. Rectification of RS distortion involves finding the camera motion for every row of RS image (row motions). Each row of the RS image is then warped using estimated row motions by taking one of the rows as reference. More than aesthetic appeal, the implications of RS rectification are critical for vision-tasks such as image registration, structure from motion (SFM), etc which perform scene inference based on geometric attributes in the captured images. 
	


Multi-frame RS rectification methods use videos \cite{liang2008analysis, ringaby2012efficient, kim2011system, grundmann} and estimate the motion across the frames using point-correspondences. The inter-frame motion helps with estimating the row motion of each RS frame, aiding the warping process to obtain distortion-free frames. Different algorithms are proposed for RS deblurring \cite{vijay_rolling_deblurring, mahesh_rolling_deblurring}, RS super-resolution \cite{abhijith_rolling_sr}, RS registration \cite{registraation1}, and change detection \cite{vijay_rolling_changedetection}.  The works in \cite{vasu2018occlusion,zhuang2017rolling} have addressed the problem of depth-aware RS rectification and again rely on multiple input frames. A differential SFM based framework is employed in \cite{zhuang2017rolling} to perform RS rectification of input images captured by a slow-moving camera. \cite{vasu2018occlusion} can handle the additional effects of occlusion that arise while capturing images using a fast-moving RS camera. Some methods have used external sensor information such as a gyroscope \cite{hee2014gyro, jia2012probabilistic,patron2015spline} to stabilize RS distortion in videos. Moreover, these methods are strongly constrained by the availability as well as reliability of external sensor data.

The afore-mentioned methods are data greedy and time-consuming except \cite{grundmann}.  Moreover, they are rendered unusable when only a single image is available. \cite{rengarajan2016bows, purkait2017rolling, lao2018robust} rely on straight lines becoming curves as a prominent effect to correct RS distortions. However, these methods are tailored for scenes that consist predominantly of straight lines and hence fail to generalize to natural images where actual curves are present in the 3D world. Moreover, they require knowledge of intrinsic camera parameters for RS rectification. 

In this paper, we address the problem of single image RS rectification using a deep neural network. The prior work to use a deep network for RS rectification for 2D scenes is \cite{rengarajan2017unrolling} wherein a neural network is trained using RS images as input, and ground truth distortion (i.e., motion) parameters as the target. Given an RS image during inference, the trained neural network predicts motion parameters corresponding to a set of key rows which is then followed by interpolation for all rows. A main drawback of this approach is that it restricts the solution space of estimated camera parameters to the ground truth parameters used during training.  Moreover, arriving at the rectified image is challenging since the association between the estimated motion parameters and the pixel position of ground truth global shutter (GS) image is unknown. \cite{rengarajan2017unrolling} attempts to solve this problem using an optimization framework as a complex post-processing step.

Recent findings in image restoration advocate that \textit{end-to-end} training performs better than decoupled or piece-wise training such as in image deblurring \cite{mathieu2015deep,tao2018scale}, ghost imaging \cite{josawang2019learning}, hyperspectral imaging \cite{josafu2020hyperspectral} and image super-resolution \cite{lim2017enhanced , kim2016accurate}. As also reiterated in \cite{yin2018fisheyerecnet}, a fisheye distortion rectification network, regressing for ground truth distortion parameters and then rectifying the distorted image gives sub-optimal performance compared to an end-to-end approach for the clean image. To this end, we propose a simple and elegant \textit{end-to-end} deep network which uses ground truth image to guide the rectification process during training. RS rectification is done in a single step during inference. 

\section{RS Image Generation and Rectification }
Rolling shutter distortion due to row-wise exposure of sensor array depends on the relative motion between camera and scene. Fig. \ref{rollingShutter} shows a scene captured using RS camera under different camera trajectories i.e., different values of $[r_x, r_y, r_z, t_x, t_y, t_z]$ where $t_{\phi}$ and $r_{\phi}$, $\phi \in \{x , y, z\}$, indicate translations along and rotations about $\phi$ axis, respectively. As observed in the figure and as also stated in \cite{rengarajan2017unrolling}, the effect of $t_y, t_z$ and $r_x $ on RS distortion is negligible as compared to the effect of $t_x, r_y$ and $r_z$. Moreover, the effect of $r_y$ can be approximated by $t_x$ for large focal length and when the movement of camera towards or away from scene is minimal. Hence, it suffices to consider only $t_x$ and $r_z$ to be essentially responsible for RS image formation. 
\newcommand{\rsi}{\ensuremath{\,\textrm{rs}}}
\newcommand{\gsi}{\ensuremath{\,\textrm{gs}}}

The GS image coordinates $(x_{\mbox{gs}}, y_{\mbox{gs}})$ are related to RS image coordinates $(x_{\mbox{rs}}, y_{\mbox{rs}})$ by 
\newcommand{\cosi}{\ensuremath{\,\textrm{cos}}}
\newcommand{\sini}{\ensuremath{\,\textrm{sin}}}

\begin{equation}\label{rsImageFormation}
    \begin{split}
        x_{\mbox{rs}} = x_{\mbox{gs}} \cdot \cosi\,(r_z(x_{\mbox{rs}})) - y_{\mbox{gs}} \cdot \sini\,(r_z(x_{\mbox{rs}})) + t_x(x_{\mbox{rs}}) \\
        y_{\mbox{rs}} = x_{\mbox{gs}} \cdot \sini\,(r_z(x_{\mbox{rs}})) + y_{\mbox{gs}} \cdot \cosi\,(r_z(x_{\mbox{rs}}))
    \end{split}
\end{equation}
where $r_z(x_{\mbox{rs}})$ and $t_x(x_{\mbox{rs}})$ are the rotation and translation motion experienced by the $x_{\mbox{rs}}^{th}$ row of RS image. The GS-RS image pairs required for training our neural network are synthesized using Eq \ref{rsImageFormation}. Given a GS image and the rotation and translation motion for every row of the RS image, the RS image can be generated using either source-to-target (S-T) or target-to-source (T-S) mapping with GS coordinates as source and RS coordinates as target. Since the motion parameters are associated with RS coordinates, S-T mapping is not employed for RS image generation. In T-S mapping, each pixel location of target RS image is multiplied with corresponding warping matrix formed by the motion parameters to yield source GS pixel location. The intensity at the resultant GS pixel coordinate is found by using bilinear interpolation and then copied to the RS pixel location.

Given an RS image and motion parameters for each row of RS image, the RS observation can be rectified akin to the process of RS image formation except that the source is now the RS image while target is the RS rectified image. In S-T mapping, each pixel location of RS image along with its row wise camera motion is substituted in Eq. (\ref{rsImageFormation}) to get RS rectified (target) pixel location. However, there is a possibility that some of the pixels in the target RS rectified image may go unfilled leaving holes in the resultant RS image. In T-S mapping, for every pixel location of RS rectified image (target), the same set of equations (i.e., Eq. (\ref{rsImageFormation})) can be used to solve for RS image (source) coordinates provided the camera motion acting on RS rectified coordinates is known.
\begin{figure*}
	\centering
	\includegraphics[height=7cm, width=13cm]{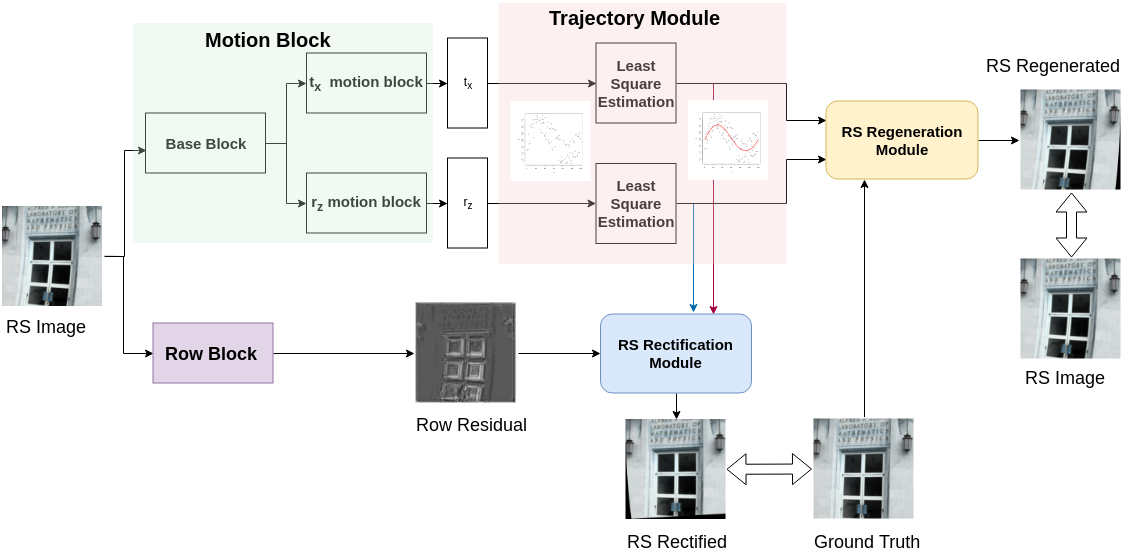}
	\caption{Proposed end-to-end deep network architecture.}
	\label{arch}
\end{figure*}
\section{Network architecture}
Our main objective is to find a direct mapping from the input RS image to the target RS rectified image. This requires estimation of row-wise camera motion parameters, and the correspondence between estimated motion parameters and target pixel locations. We achieve the above in a principled manner as follows. We propose to use image features of input RS image for estimating camera motion parameters and devise a mapping function to relate estimated motion parameters with pixel locations of target image.

Our network architecture (Fig. \ref{arch}) consists of five basic modules: motion block, trajectory module, row block, RS regeneration module and RS rectification module. The motion block predicts camera motion parameters ($t_x$ and $r_z$) for each row of the input RS image. The trajectory module ensures that the estimated camera motion parameters follow a smooth and continuous curve as we traverse the rows in the RS image, in compliance with real-life camera trajectories. For each pixel location in the target image, the corresponding camera motion is found using the row block. The output of row block and trajectory module are used by the RS rectification module to warp the input RS image to get the RS rectified image. For faster convergence during training and to better-condition the optimisation process, we also employ an RS regeneration module which takes motion parameters from the trajectory module and warps the GS image to estimate the given (input) RS image. A detailed discussion of each of the module follows.

\textbf{Motion block} This consists of a base block followed by $t_{x}$(translation) and $r_{z}$(rotation) blocks, respectively. The base block extracts features from the input RS image which are then used to find row wise translation and rotation motion of the input image. Thus, the motion block takes input color image of size $r \times r \times 3$ and outputs two 1D vectors of length $r$ indicating rotation and translation motion parameters for each row of the input RS image. The base block is designed using three convolutional layers. Both translation and rotation blocks, which take the output of the base network, are designed using three convolutional layers followed by two fully connected (FC) layers. The final FC layer is of dimension [$r$,1] reflecting the motion for every row of the input RS image. Each convolutional layer is followed by batch normalization.

\textbf{Row block} As discussed in the second section, every pixel coordinate in the GS (target) image is substituted in Eq. (\ref{rsImageFormation}) along with the corresponding camera motion to get the RS (source) image coordinate. However, camera motion acting on each GS pixel to form the RS image is known only to the extent that it is one of the motions experienced by the rows of RS image. Specifically, for a given input RS image, all the pixels in a row stem from a single camera motion and the motion will generally be different for each row. In contrast, pixels in a row of the GS image need not be influenced by a single motion.
The ambiguity of which motion to associate to a pixel in target image was addressed in \cite{rengarajan2017unrolling} as a post-processing step, which is a complicated exercise and implicitly constrains the estimated motion parameters. 
 
 We propose to use a deep network to solve this issue thus rendering our network \textit{end-to-end}. The row block takes an image of size $r\times r\times 3$ and outputs a matrix of dimension $r\times r$, with each location indicating which row motion of RS image must be considered from the estimated motion parameters for rectification. In real camera trajectory, camera motion is typically smooth. Consequently, the output of row block at each coordinate location can be expected to be close to its corresponding row number. Hence, for both stability and faster convergence, we learn to solve only for the residual (motivated by \cite{he2016deep,kim2016accurate}).  The residual image is an $r \times  r$  matrix and is the output of the row block. This image is added to another matrix (say $A$) which is initialized with its own row number i.e., $A(i,j) = i$ for the reason stated above. The resultant matrix values indicate which camera motion is to be considered from the estimated motion parameters for rectification of  RS image. From now, we refer to output of row block as the sum of learnt residual with a matrix initialized with its row number at each coordinate position. The row block consists of five convolutional layers with each layer followed by batch normalization and an activation function. Use of three layers for base block and a total of 6 convolution layers for motion block is partly motivated by \cite{rengarajan2017unrolling} (it uses five convolution layers for motion estimation). Since, the objective of row block (finding residual for)  is comparatively less complex compared to motion block we used only 3 layers.
 
\begin{figure*}

\begin{tabular}{c c c c c c c}
\centering

\begin{subfigure}{0.15\linewidth}
\includegraphics[width=\linewidth]{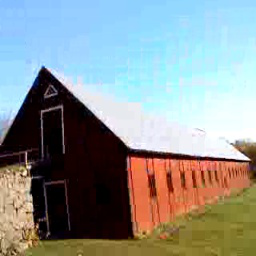}
\caption*{Input RS Frames}
\end{subfigure} &
\hspace{-0.3cm}
\begin{subfigure}{0.15\linewidth}
\includegraphics[width=\linewidth]{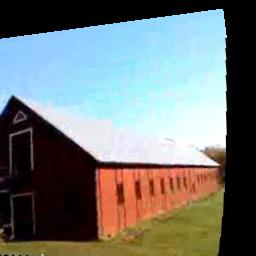}
\caption*{\cite{purkait2017rolling}, $|R_F|$=29.0}
\end{subfigure} &
\hspace{-0.3cm}
\begin{subfigure}{0.15\linewidth}
\includegraphics[width=\linewidth]{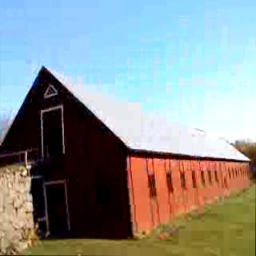}
\caption*{\cite{rengarajan2016bows}, $|R_F|$=35.7}
\end{subfigure} &
\hspace{-0.3cm}
\begin{subfigure}{0.15\linewidth}
\includegraphics[width=\linewidth]{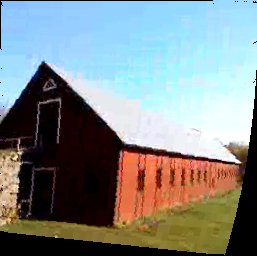}
\caption*{\cite{lao2018robust}, $|R_F|$=40.7}
\end{subfigure} &
\hspace{-0.3cm}

\begin{subfigure}{0.15\linewidth}
\includegraphics[width=\linewidth]{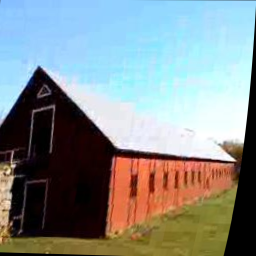}
\caption*{\cite{rengarajan2017unrolling}, $|R_F|$=41.7}
\end{subfigure} &
\hspace{-0.3cm}

\begin{subfigure}{0.15\linewidth}
\includegraphics[width=\linewidth]{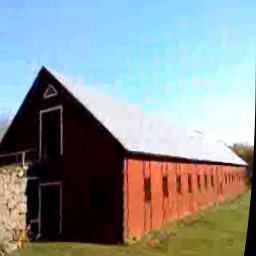}
\caption*{Ours,$|R_F|$=45.92}
\end{subfigure}

\end{tabular}
\caption{Comparison of $|R_F|$ value with different algorithms on a real video.}
\label{rfImages}
\end{figure*}

\subsection{Loss functions}
Given an input RS image, using the trajectory module, row block and RS rectification module, an input image can be rectified to give an RS distortion-free or rectified image. We employ different loss functions to enable the network to learn the rectification process. 

The first loss is the mean squared error (MSE) between rectified RS image and ground truth image but with a modification. In the rectified RS image, it is possible that certain regions in the boundary are not recovered (when compared with GS image) since these regions were not present in the RS image itself due to camera motion. This can be noticed in Fig. 5 (third row) where the building has been rectified but there are regions on the boundary where the rectification algorithm could not retrieve pixel values as they were not present in the original RS image. To account for this effect, we used a visibility aware MSE loss where MSE is considered between two pixels only if the intensity in at least one of the channels in the rectified RS image is non-zero. Let $I_{\mbox{rs}}, I_{\mbox{gs}}, I_{\mbox{rs\_rec}}$ be input RS, ground truth GS, and RS rectified image, respectively. Then, we define mask $M_{\mbox{rs\_rec}}$, such that  
\newcommand{\inch}{\ensuremath{\,\textrm{otherwise}}}
\begin{equation*}
M_{\mbox{rs\_rec}}(i,j) = \begin{cases} 0 & \mbox{if} \sum \limits_{k=1}^3 I_{\mbox{rs\_rec}}(i,j,k) = 0\\ 1 & \inch \end{cases}
\end{equation*}
where $k$ indicates color channel in the RGB image. The error between GS and RS rectified image can be written as  
\begin{equation*}
L_{\mbox{rs\_rec\_MSE}} = || I_{\mbox{rs\_rec}} - M_{\mbox{rs\_rec}} \otimes I_{\mbox{gs}}||^2_2
\end{equation*}
where $\otimes$ refers to point-wise multiplication.

The second loss that we devise is based on the error between the given RS image and the GS image distorted by estimated motion parameters. To account for holes in the boundary, we again define mask $M_{\mbox{rs\_reg}}(i,j)$ such that 
\begin{equation*}
M_{\mbox{rs\_reg}}(i,j) = \begin{cases} 0 & \mbox{if} \sum \limits_{k=1}^3 I_{\mbox{rs\_reg}}(i,j,k) = 0\\ 1 & \inch \end{cases}
\end{equation*}
where $I_{\mbox{rs\_reg}}$ is the image obtained by applying estimated motion parameters on the GS image. The error between the RS image and the RS regenerated image is given by 
\begin{equation*}
L_{\mbox{rs\_reg\_MSE}} = || I_{\mbox{rs\_reg}} - M_{\mbox{rs\_reg}} \otimes I_{\mbox{rs}}||^2_2
\end{equation*}
Since edges play a very important role in RS rectification, we also compare Sobel edges of RS rectified and RS regenerated images with ground truth GS and input RS images, respectively. Let the Sobel operation be represented as $E(.)$. Then the edge losses for regeneration phase and rectification phase can be formulated as
\begin{equation*}
L_{\mbox{rs\_rec\_edge}} = || E(I_{\mbox{rs\_rec}}) - M_{\mbox{rs\_rec}} \otimes E(I_{\mbox{gs}})||^2_2
\end{equation*}
\begin{equation*}
L_{\mbox{rs\_reg\_edge}} = || E(I_{\mbox{rs\_reg}}) - M_{\mbox{rs\_reg}} \otimes E(I_{\mbox{rs}})||^2_2
\end{equation*}
The overall loss function (please refer to Appendix for back propagation equations w.r.t different loss functions) of our network is a combination of the afore-mentioned loss functions and is given by 
\begin{multline}
L_{total} = \lambda_1L_{\mbox{rs\_rec\_MSE}} + \lambda_2 L_{\mbox{rs\_reg\_MSE}}
+ \lambda_3 L_{\mbox{rs\_rec\_edge}} + \lambda_4 L_{\mbox{rs\_reg\_edge}}
\label{total_sum_loss}
\end{multline}

\begin{figure*}
\hspace{-0.8cm}
\begin{tabular}{c c c c c c c}
\centering
\begin{subfigure}{0.15\linewidth}
\includegraphics[width=\linewidth]{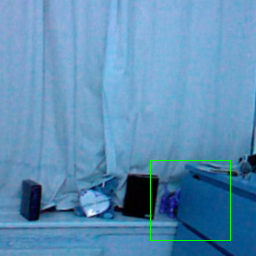}
\end{subfigure} &
\hspace{-0.45cm}

\begin{subfigure}{0.15\linewidth}
\includegraphics[width=\linewidth]{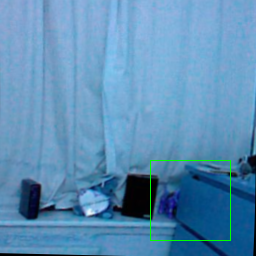}
\end{subfigure} &
\hspace{-0.45cm}

\begin{subfigure}{0.15\linewidth}
\includegraphics[width=\linewidth]{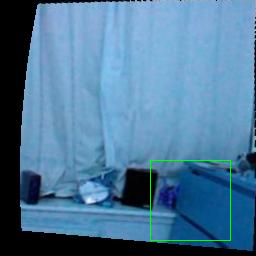}
\end{subfigure} &
\hspace{-0.45cm}

\begin{subfigure}{0.15\linewidth}
\includegraphics[width=\linewidth]{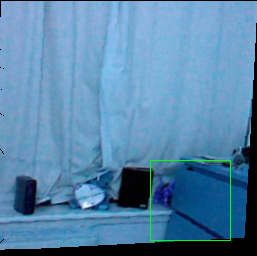}
\end{subfigure} &
\hspace{-0.45cm}

\begin{subfigure}{0.15\linewidth}
\includegraphics[width=\linewidth]{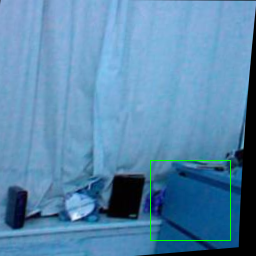}
\end{subfigure} &
\hspace{-0.45cm}

\begin{subfigure}{0.15\linewidth}
\includegraphics[width=\linewidth]{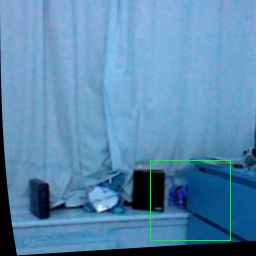}
\end{subfigure} &
\hspace{-0.45cm}

\begin{subfigure}{0.15\linewidth}
\includegraphics[width=\linewidth]{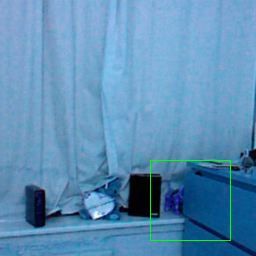}
\end{subfigure}
\\
\begin{subfigure}{0.15\linewidth}
\includegraphics[width=\linewidth]{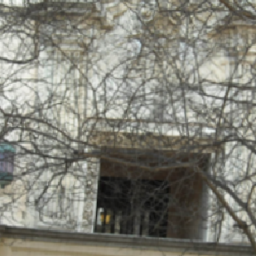}
\end{subfigure} &
\hspace{-0.45cm}

\begin{subfigure}{0.15\linewidth}
\includegraphics[width=\linewidth]{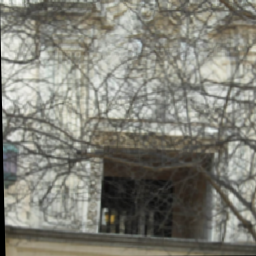}
\end{subfigure} &
\hspace{-0.45cm}

\begin{subfigure}{0.15\linewidth}
\includegraphics[width=\linewidth]{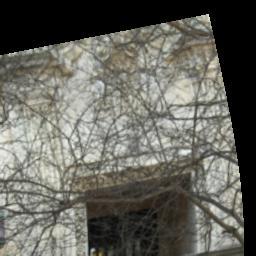}
\end{subfigure} &
\hspace{-0.45cm}

\begin{subfigure}{0.15\linewidth}
\includegraphics[width=\linewidth]{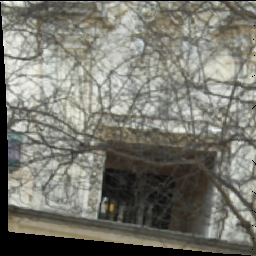}
\end{subfigure} &
\hspace{-0.45cm}

\begin{subfigure}{0.15\linewidth}
\includegraphics[width=\linewidth]{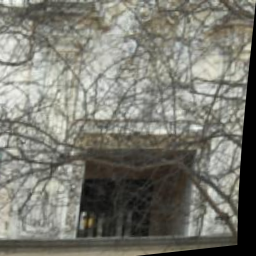}
\end{subfigure} &
\hspace{-0.45cm}

\begin{subfigure}{0.15\linewidth}
\includegraphics[width=\linewidth]{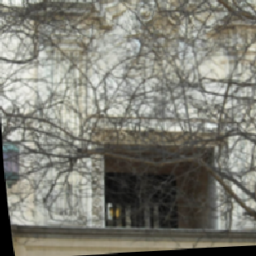}
\end{subfigure} &
\hspace{-0.45cm}

\begin{subfigure}{0.15\linewidth}
\includegraphics[width=\linewidth]{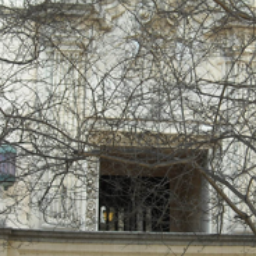}
\end{subfigure} \\ \\


\begin{subfigure}{0.15\linewidth}
\includegraphics[width=\linewidth]{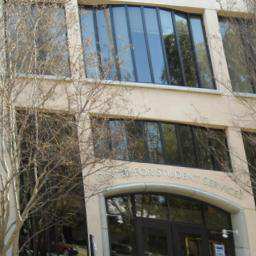}
\caption*{Input RS image}
\end{subfigure} &
\hspace{-0.45cm}

\begin{subfigure}{0.15\linewidth}
\includegraphics[width=\linewidth]{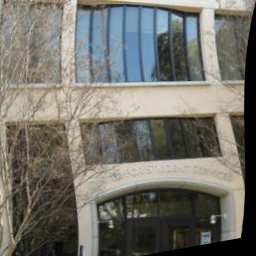}
\caption*{\cite{rengarajan2016bows}}
\end{subfigure} &
\hspace{-0.45cm}

\begin{subfigure}{0.15\linewidth}
\includegraphics[width=\linewidth]{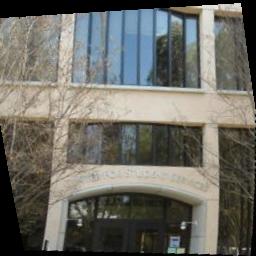}
\caption*{\cite{purkait2017rolling}}
\end{subfigure} &
\hspace{-0.45cm}
\begin{subfigure}{0.15\linewidth}
\includegraphics[width=\linewidth]{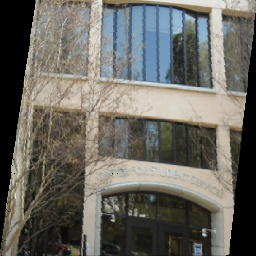}
\caption*{\cite{lao2018robust}}
\end{subfigure} &
\hspace{-0.45cm}

\begin{subfigure}{0.15\linewidth}
\includegraphics[width=\linewidth]{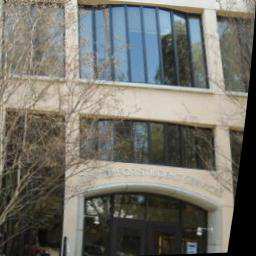}
\caption*{\cite{rengarajan2017unrolling} }
\end{subfigure} &
\hspace{-0.45cm}

\begin{subfigure}{0.15\linewidth}
\includegraphics[width=\linewidth]{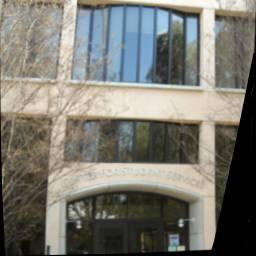}
\caption*{Ours}
\end{subfigure} &
\hspace{-0.45cm}

\begin{subfigure}{0.15\linewidth}
\includegraphics[width=\linewidth]{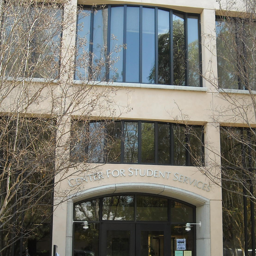}
\caption*{Ground truth}
\end{subfigure} \\ \\
\end{tabular}
\caption{Visual comparisons on synthetic examples with different RS rectification methods.}
\label{allComparison}
\end{figure*}

\section{Experiments}
This section is arranged as follows: (i) dataset generation, (ii)  implementation details, (iii) competing methods, (iv) quantitative analysis, and (v) visual results.
\subsection{Dataset generation }
To train our network, we used images of buildings but tested on buildings, as well as real-world RS images (having atleast few real-world straight lines). In this section, we explain the synthesis of camera motion and generation of RS images for training. Since fully connected (FC) layers are present in our network, we used images of constant size (256x256) for both training and testing.\\
\textbf{Camera motion and Training dataset:} Because it is difficult to capture real GS-RS pairs, following \cite{rengarajan2017unrolling} we synthesized camera motion using a second-degree polynomial for generating the RS images. We used the Buildings dataset from \cite{xiao2010sun,shao2003zubud} with a total of 440 clean images cropped to a size of 256x256. Out of those, we randomly chose 400 images and each image is distorted using 200 synthesized camera motions resulting in 80K images for training. The remaining 40 images are used in the test dataset. In order to ensure that there are no missing parts in the boundaries of generated RS images, we increased the size of each image to 356x356, applied RS distortions, and then cropped them back to 256x256.\\
\subsection{Implementation details and competing methods}
To stabilize training and mitigate the ill-conditioness during the initial steps, we trained our network to regress for only ground truth motion parameters using 50 images from the training dataset for 5 epochs.
Then the network is trained using Eq. 2 as our loss function with full size training dataset. We used TensorFlow for both training and testing with following options: ADAM optimizer to minimize the loss function, momentum values with $\beta_1$ = 0.9 , $\beta_2$ = 0.99 and with a learning rate of 0.001. The weights of different cost functions are set as $\lambda_1$ = $\lambda_2$ = 1 and $\lambda_3$ = $\lambda_4$ = 0.5. We compared our method with state-of-the-art single image RS rectification methods \cite{rengarajan2017unrolling, lao2018robust, purkait2017rolling, rengarajan2016bows}. Note that non-learning based methods \cite{rengarajan2016bows, lao2018robust,purkait2017rolling} require intrinsic camera parameters while ours and \cite{rengarajan2017unrolling} do not. We gave our set of RS images for comparison to the respective authors and obtained the results from them.

\subsection{Visual comparisons}
We give results on the test dataset and RS images captured using a hand-held mobile camera. Fig. \ref{allComparison} depicts qualitative comparisons with competing methods. 
 The RS image in the first row is an Indoor scene image (\cite{silberman2012indoor}) affected by a real-life camera trajectory (\cite{kohler2012recording}). Because the Manhattan world assumption is not satisfied and due to the presence of influential outliers in the background (cloth), \cite{lao2018robust, purkait2017rolling, rengarajan2016bows} is not able to rectify the image properly. Our rectification result is better than that of \cite{rengarajan2017unrolling} since the solution space of estimated motion parameters is not skewed, unlike \cite{rengarajan2017unrolling}. 
The RS images in the second and third (taken from \cite{lao2018robust}) row, also part of test dataset, are affected by complex real-life camera motion which is evident from the RS distortions. Since strong outliers are present in these images in the form of branches, \cite{lao2018robust, purkait2017rolling, rengarajan2016bows} which depend on the detection of curves for estimation of camera motion fail to rectify the image. Due to restrictions on estimated camera motion estimation, \cite{rengarajan2017unrolling} is unable to properly rectify the images in comparison to ours.

\begin{figure}[t]
	\centering
	\begin{subfigure}[b]{0.25\textwidth}
		\includegraphics[width=0.95\linewidth]{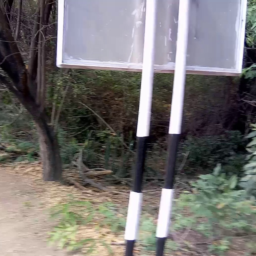}
		\caption{}
		
	\end{subfigure}%
	\begin{subfigure}[b]{0.25\textwidth}
		\includegraphics[width=0.95\linewidth]{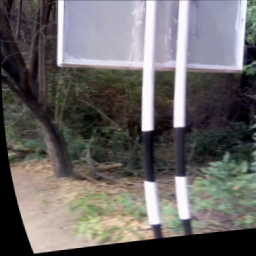}
		\caption{}
		
	\end{subfigure}%
	\begin{subfigure}[b]{0.25\textwidth}
		\includegraphics[width=0.95\linewidth]{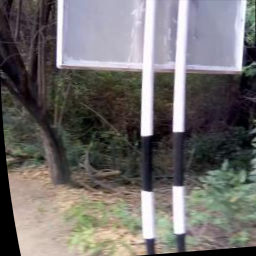}
		\caption{}
	
	\end{subfigure}%
	\begin{subfigure}[b]{0.25\textwidth}
		\includegraphics[width=0.95\linewidth]{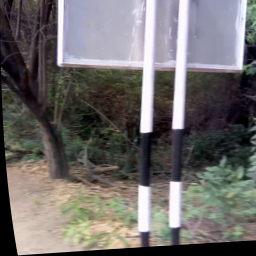}
		\caption{}	
	\end{subfigure}%

\caption{Ablation study. (a) Input RS real image. (b) Result of our model regressed for ground truth motion parameters. (c) Result of \cite{rengarajan2017unrolling}. (d) Our result.}
\label{ablation_studies2}
\end{figure}

Refined and complete version of this work appeared in JOSA 2020

\bibliographystyle{plain}
\bibliography{egbib}

\begin{thebibliography}{10}

\bibitem{josafu2020hyperspectral}
Hao Fu, Liheng Bian, Xianbin Cao, and Jun Zhang.
\newblock Hyperspectral imaging from a raw mosaic image with end-to-end
  learning.
\newblock {\em Optics Express}, 28(1):314--324, 2020.

\bibitem{grundmann}
Matthias Grundmann, Vivek Kwatra, Daniel Castro, and Irfan Essa.
\newblock Calibration-free rolling shutter removal.
\newblock In {\em 2012 IEEE international conference on computational
  photography (ICCP)}, pages 1--8. IEEE, 2012.

\bibitem{he2016deep}
Kaiming He, Xiangyu Zhang, Shaoqing Ren, and Jian Sun.
\newblock Deep residual learning for image recognition.
\newblock In {\em Proceedings of the IEEE conference on computer vision and
  pattern recognition}, pages 770--778, 2016.

\bibitem{hee2014gyro}
Sung Hee~Park and Marc Levoy.
\newblock Gyro-based multi-image deconvolution for removing handshake blur.
\newblock In {\em Proceedings of the IEEE Conference on Computer Vision and
  Pattern Recognition}, pages 3366--3373, 2014.

\bibitem{jia2012probabilistic}
Chao Jia and Brian~L Evans.
\newblock Probabilistic 3-d motion estimation for rolling shutter video
  rectification from visual and inertial measurements.
\newblock In {\em 2012 IEEE 14th International Workshop on Multimedia Signal
  Processing (MMSP)}, pages 203--208. IEEE, 2012.

\bibitem{kim2016accurate}
Jiwon Kim, Jung Kwon~Lee, and Kyoung Mu~Lee.
\newblock Accurate image super-resolution using very deep convolutional
  networks.
\newblock In {\em Proceedings of the IEEE conference on computer vision and
  pattern recognition}, pages 1646--1654, 2016.

\bibitem{kim2011system}
Young-Geun Kim, Venkata~Ravisankar Jayanthi, and In-So Kweon.
\newblock System-on-chip solution of video stabilization for cmos image sensors
  in hand-held devices.
\newblock {\em IEEE transactions on circuits and systems for video technology},
  21(10):1401--1414, 2011.

\bibitem{kohler2012recording}
Rolf K{\"o}hler, Michael Hirsch, Betty Mohler, Bernhard Sch{\"o}lkopf, and
  Stefan Harmeling.
\newblock Recording and playback of camera shake: Benchmarking blind
  deconvolution with a real-world database.
\newblock In {\em European conference on computer vision}, pages 27--40.
  Springer, 2012.

\bibitem{lao2018robust}
Yizhen Lao and Omar Ait-Aider.
\newblock A robust method for strong rolling shutter effects correction using
  lines with automatic feature selection.
\newblock In {\em Proceedings of the IEEE Conference on Computer Vision and
  Pattern Recognition}, pages 4795--4803, 2018.

\bibitem{liang2008analysis}
Chia-Kai Liang, Li-Wen Chang, and Homer~H Chen.
\newblock Analysis and compensation of rolling shutter effect.
\newblock {\em IEEE Transactions on Image Processing}, 17(8):1323--1330, 2008.

\bibitem{lim2017enhanced}
Bee Lim, Sanghyun Son, Heewon Kim, Seungjun Nah, and Kyoung Mu~Lee.
\newblock Enhanced deep residual networks for single image super-resolution.
\newblock In {\em Proceedings of the IEEE Conference on Computer Vision and
  Pattern Recognition Workshops}, pages 136--144, 2017.

\bibitem{mathieu2015deep}
Michael Mathieu, Camille Couprie, and Yann LeCun.
\newblock Deep multi-scale video prediction beyond mean square error.
\newblock {\em arXiv preprint arXiv:1511.05440}, 2015.

\bibitem{mahesh_rolling_deblurring}
Mahesh~MR Mohan, AN~Rajagopalan, and Gunasekaran Seetharaman.
\newblock Going unconstrained with rolling shutter deblurring.
\newblock In {\em Proceedings of the IEEE International Conference on Computer
  Vision}, pages 4010--4018, 2017.

\bibitem{patron2015spline}
Alonso Patron-Perez, Steven Lovegrove, and Gabe Sibley.
\newblock A spline-based trajectory representation for sensor fusion and
  rolling shutter cameras.
\newblock {\em International Journal of Computer Vision}, 113(3):208--219,
  2015.

\bibitem{vijay_rolling_changedetection}
Vijay Rengarajan~Angarai Pichaikuppan, Rajagopalan~Ambasamudram Narayanan, and
  Aravind Rangarajan.
\newblock Change detection in the presence of motion blur and rolling shutter
  effect.
\newblock In {\em European Conference on Computer Vision}, pages 123--137.
  Springer, 2014.

\bibitem{abhijith_rolling_sr}
Abhijith Punnappurath, Vijay Rengarajan, and AN~Rajagopalan.
\newblock Rolling shutter super-resolution.
\newblock In {\em Proceedings of the IEEE International Conference on Computer
  Vision}, pages 558--566, 2015.

\bibitem{purkait2017rolling}
Pulak Purkait, Christopher Zach, and Ales Leonardis.
\newblock Rolling shutter correction in manhattan world.
\newblock In {\em Proceedings of the IEEE International Conference on Computer
  Vision}, pages 882--890, 2017.

\bibitem{rengarajan2017unrolling}
Vijay Rengarajan, Yogesh Balaji, and AN~Rajagopalan.
\newblock Unrolling the shutter: Cnn to correct motion distortions.
\newblock In {\em Proceedings of the IEEE Conference on Computer Vision and
  Pattern Recognition}, pages 2291--2299, 2017.

\bibitem{rengarajan2016bows}
Vijay Rengarajan, Ambasamudram~N Rajagopalan, and Rangarajan Aravind.
\newblock From bows to arrows: Rolling shutter rectification of urban scenes.
\newblock In {\em Proceedings of the IEEE Conference on Computer Vision and
  Pattern Recognition}, pages 2773--2781, 2016.

\bibitem{vijay_rolling_deblurring}
Vijay Rengarajan, Ambasamudram~Narayanan Rajagopalan, Rangarajan Aravind, and
  Guna Seetharaman.
\newblock Image registration and change detection under rolling shutter motion
  blur.
\newblock {\em IEEE Transactions on Pattern Analysis and Machine Intelligence},
  39(10):1959--1972, 2017.

\bibitem{ringaby2012efficient}
Erik Ringaby and Per-Erik Forss{\'e}n.
\newblock Efficient video rectification and stabilisation for cell-phones.
\newblock {\em International Journal of Computer Vision}, 96(3):335--352, 2012.

\bibitem{shao2003zubud}
Hao Shao, Tom{\'a}{\v{s}} Svoboda, and Luc Van~Gool.
\newblock Zubud-zurich buildings database for image based recognition.
\newblock {\em Computer Vision Lab, Swiss Federal Institute of Technology,
  Switzerland, Tech. Rep}, 260(20):6--8, 2003.

\bibitem{silberman2012indoor}
Nathan Silberman, Derek Hoiem, Pushmeet Kohli, and Rob Fergus.
\newblock Indoor segmentation and support inference from rgbd images.
\newblock In {\em European Conference on Computer Vision}, pages 746--760.
  Springer, 2012.

\bibitem{tao2018scale}
Xin Tao, Hongyun Gao, Xiaoyong Shen, Jue Wang, and Jiaya Jia.
\newblock Scale-recurrent network for deep image deblurring.
\newblock In {\em Proceedings of the IEEE Conference on Computer Vision and
  Pattern Recognition}, pages 8174--8182, 2018.

\bibitem{vasu2018occlusion}
Subeesh Vasu, Mahesh~MR Mohan, and AN~Rajagopalan.
\newblock Occlusion-aware rolling shutter rectification of 3d scenes.
\newblock In {\em Proceedings of the IEEE Conference on Computer Vision and
  Pattern Recognition}, pages 636--645, 2018.

\bibitem{registraation1}
Subeesh Vasu, Ambasamudram~Narayanan Rajagopalan, and Guna Seetharaman.
\newblock Camera shutter-independent registration and rectification.
\newblock {\em IEEE Transactions on Image Processing}, 27(4):1901--1913, 2018.

\bibitem{josawang2019learning}
Fei Wang, Hao Wang, Haichao Wang, Guowei Li, and Guohai Situ.
\newblock Learning from simulation: An end-to-end deep-learning approach for
  computational ghost imaging.
\newblock {\em Optics express}, 27(18):25560--25572, 2019.

\bibitem{xiao2010sun}
Jianxiong Xiao, James Hays, Krista~A Ehinger, Aude Oliva, and Antonio Torralba.
\newblock Sun database: Large-scale scene recognition from abbey to zoo.
\newblock In {\em 2010 IEEE Computer Society Conference on Computer Vision and
  Pattern Recognition}, pages 3485--3492. IEEE, 2010.

\bibitem{yin2018fisheyerecnet}
Xiaoqing Yin, Xinchao Wang, Jun Yu, Maojun Zhang, Pascal Fua, and Dacheng Tao.
\newblock Fisheyerecnet: A multi-context collaborative deep network for fisheye
  image rectification.
\newblock In {\em Proceedings of the European Conference on Computer Vision
  (ECCV)}, pages 469--484, 2018.

\bibitem{zhuang2017rolling}
Bingbing Zhuang, Loong-Fah Cheong, and Gim Hee~Lee.
\newblock Rolling-shutter-aware differential sfm and image rectification.
\newblock In {\em Proceedings of the IEEE International Conference on Computer
  Vision}, pages 948--956, 2017.

\end{thebibliography}
\end{document}